\documentclass[conference]{IEEEtran}
\IEEEoverridecommandlockouts

\makeatletter
\def\ps@IEEEtitlepagestyle{%
  \def\@oddfoot{\mycopyrightnotice}%
  \def\@oddhead{\hbox{}\@IEEEheaderstyle\leftmark\hfil\thepage}\relax
  \def\@evenhead{\@IEEEheaderstyle\thepage\hfil\leftmark\hbox{}}\relax
  \def\@evenfoot{}%
}
\def\mycopyrightnotice{%
  \begin{minipage}{\textwidth}
  \centering \scriptsize
  979-8-3315-5397-5/25/\$31.00 ©  
  \textbf{2025 IEEE International Conference on Future Machine Learning and Data Science (FMLDS 2025)}.
  \end{minipage}
}
\makeatother

\usepackage{array}
\usepackage[caption=false,font=normalsize,labelfont=sf,textfont=sf]{subfig}
\usepackage[inline]{enumitem}
\usepackage{textcomp}
\usepackage{url}
\usepackage{verbatim}
\usepackage{graphicx}

\usepackage{xcolor}

\usepackage{tabularx}
\usepackage{booktabs}
\newcommand*{\mystrut}{\rule{0pt}{3.5ex}}
\usepackage[table]{xcolor}

\usepackage{makecell} 

\usepackage{cite}
\usepackage{amsmath,amssymb,amsfonts}
\usepackage{algorithmic}
\usepackage{graphicx}
\usepackage{textcomp}

\usepackage{fancyhdr}
\usepackage[vlined]{algorithm2e}
\usepackage{dblfloatfix}
\usepackage{booktabs} 
\usepackage{makecell} 
\usepackage{colortbl} 

\usepackage{multirow}
\usepackage{caption}
\usepackage{float}
\usepackage{capt-of}
\usepackage{varwidth}
\usepackage{setspace, etoolbox}

\usepackage{pifont}

\usepackage{pdflscape} 

\newcommand\M{\mathcal{M}}
\newcommand\I{\mathcal{I}}
\newcommand\B{\mathcal{B}}
\newcommand\G{\mathcal{G}}
\newcommand\Z{\mathcal{Z}}
\newcommand\Q{\mathcal{Q}}
\newcommand\N{\mathcal{N}}
\newcommand\C{\mathcal{C}}
\newcommand\UCB{\mathcal{UCB}}

\DeclareMathOperator*{\argmax}{arg\,max}

\SetCommentSty{mycommfont}

\usepackage{amsmath}
\usepackage{wrapfig}

\usepackage{setspace, etoolbox}
\usepackage{subcaption}

\usepackage{amssymb}

\newcommand\D{\mathcal{D}}

\SetKwComment{Comment}{$\triangleright$\ }{}
\SetKwInput{KwDefinition}{\color{cyan}Function}
\SetKw{KwResult} {\color{cyan}Output:}

\usepackage[square,sort,comma,numbers]{natbib}
\newcommand{\citean}[1]{\citeauthor{#1}~\cite{#1}}

\def\BibTeX{{\rm B\kern-.05em{\sc i\kern-.025em b}\kern-.08em
    T\kern-.1667em\lower.7ex\hbox{E}\kern-.125emX}}
\usepackage{balance}

\usepackage[pdfstartview=XYZ,
bookmarks=true,
colorlinks=true,
linkcolor=blue,
urlcolor=blue,
citecolor=blue,
pdftex,
bookmarks=true,
linktocpage=true, 
hyperindex=true
]{hyperref}

\usepackage{orcidlink}
\begin{document}

\makeatletter
    \newcommand{\linebreakand}{%
      \end{@IEEEauthorhalign}
      \hfill\mbox{}\par
      \mbox{}\hfill\begin{@IEEEauthorhalign}
    }
    \makeatother
    
\title{Policy-Driven Transfer Learning in Resource-Limited Animal Monitoring
}
\author{\IEEEauthorblockN{Nisha Pillai\IEEEauthorrefmark{1}}
\IEEEauthorblockA{\textit{Mississippi State University} }
\and
\IEEEauthorblockN{Aditi Virupakshaiah\IEEEauthorrefmark{2}}
\IEEEauthorblockA{\textit{Mississippi State University}}
\and 
\IEEEauthorblockN{Harrison W. Smith\IEEEauthorrefmark{3}}
\IEEEauthorblockA{\textit{University of Arkansas}}
\and
\IEEEauthorblockN{Amanda J. Ashworth\IEEEauthorrefmark{4}}
\IEEEauthorblockA{\textit{USDA-ARS} }
\linebreakand
\IEEEauthorblockN{Prasanna Gowda\IEEEauthorrefmark{5}}
\IEEEauthorblockA{\textit{USDA-ARS}}
\and
\IEEEauthorblockN{Phillip R. Owens\IEEEauthorrefmark{6}}
\IEEEauthorblockA{\textit{USDA-ARS}}
\and
\IEEEauthorblockN{Adam R. Rivers\IEEEauthorrefmark{7}}
\IEEEauthorblockA{\textit{USDA-ARS}}
\and
\IEEEauthorblockN{Bindu Nanduri\IEEEauthorrefmark{8}}
\IEEEauthorblockA{\textit{Mississippi State University}}
\and
\IEEEauthorblockN{Mahalingam Ramkumar\IEEEauthorrefmark{9}}
\IEEEauthorblockA{\textit{Mississippi State University}}

\linebreakand
\IEEEauthorblockA{Email:\{\IEEEauthorrefmark{1}pillai, \IEEEauthorrefmark{9}ramkumar\}@cse.msstate.edu, \IEEEauthorrefmark{2}av807@msstate.edu, \IEEEauthorrefmark{3}hws001@uark.edu, \IEEEauthorrefmark{8}bnanduri@cvm.msstate.edu}
\linebreakand
\IEEEauthorblockA{\{\IEEEauthorrefmark{4}amanda.ashworth, \IEEEauthorrefmark{5}prasanna.gowda,\IEEEauthorrefmark{6}phillip.owens, \IEEEauthorrefmark{7}adam.rivers\}@usda.gov}

}


    
    
    
\maketitle

\begin{abstract}
Animal health monitoring and population management are critical aspects of wildlife conservation and livestock management that increasingly rely on automated detection and tracking systems. While Unmanned Aerial Vehicle (UAV) based systems combined with computer vision offer promising solutions for non-invasive animal monitoring across challenging terrains, limited availability of labeled training data remains an obstacle in developing effective deep learning (DL) models for these applications. Transfer learning has emerged as a potential solution, allowing models trained on large datasets to be adapted for resource-limited scenarios such as those with limited data. However, the vast landscape of pre-trained neural network architectures makes it challenging to select optimal models, particularly for researchers new to the field. In this paper, we propose a reinforcement learning (RL)-based transfer learning framework that employs an upper confidence bound (UCB) algorithm to automatically select the most suitable pre-trained model for animal detection tasks. Our approach systematically evaluates and ranks candidate models based on their performance, streamlining the model selection process. Experimental results demonstrate that our framework achieves a higher detection rate while requiring significantly less computational time compared to traditional methods.

\end{abstract}


\section{Introduction}

Animal counting is a fundamental yet crucial aspect of livestock monitoring and animal health management~\cite{ norouzzadeh2018automatically}. Accurate population counts enable researchers and conservationists to track species abundance, assess ecosystem health, and identify concerning trends in animal populations that may indicate environmental stressors or disease outbreaks~\cite{green2005complexity}. Regular counting helps establish baseline population data~\cite{kuhl2008best} and detect abnormal fluctuations that could signal health issues affecting a particular species or habitat. The advent of automated counting technologies, including drone surveys and AI-powered image analysis, has greatly enhanced our ability to conduct frequent and accurate population assessments, leading to more timely interventions when health issues arise and better-informed conservation strategies~\cite{buchelt2024exploring}.

The automated counting of animals through computer vision represents a critical advancement in wildlife monitoring and conservation efforts~\cite{kumar2023comprehensive, chen2023using}, yet it presents significant technical and computational challenges. Traditional approaches rely on manual counting methods that are time-consuming and prone to human error, making automated solutions increasingly essential~\cite{petso2022review}. The development of such systems requires a comprehensive pipeline that begins with extensive data collection using field cameras, drones, and manual photography to capture animals in diverse environmental conditions~\cite{besson2022towards}. This data undergoes rigorous preprocessing, including image normalization, dimensional standardization, and augmentation techniques to enhance dataset robustness. The core methodology employs deep learning-based object detection and tracking algorithms, typically leveraging transfer learning from pre-trained models. In this research, we use these models without fine-tuning on our test dataset to compensate for limited computational resources and training data~\cite{hosain2024synchronizing}. However, the implementation of these systems presents unique challenges for bioinformaticians and machine learning practitioners, particularly in resource-constrained environments where access to high-performance computing infrastructure is limited. These constraints necessitate careful optimization of model architectures and training strategies to balance computational efficiency with detection accuracy, while still maintaining the ability to handle complex scenarios such as occlusion, varying lighting conditions, and animal movement patterns.

The evolution of object detection has yielded an extensive array of algorithmic choices, moving beyond conventional approaches like Faster R-CNN and SSD to embrace modern architectures such as YOLO, EfficientDet, and transformer-based systems~\cite{pal2021deep, khanam2024comprehensive}. While this diversity offers tremendous potential for specialized applications, it presents a substantial challenge in selecting the optimal algorithm for specific use cases like animal detection, welfare assessment, and counting. The conventional approach of exhaustively evaluating each model on large datasets is computationally prohibitive and time-consuming, often requiring weeks or months of experimentation to determine the most effective solution. To address this challenge, we propose a reinforcement learning-based approach for algorithm selection that significantly reduces the computational overhead of model evaluation. Our method employs a reward system that adaptively learns the effectiveness of different models based on their performance on data subsets, eliminating the need for comprehensive testing of each algorithm. Furthermore, we propose an upper confidence bound (UCB)-based selection strategy that maintains a crucial balance between exploiting high-performing algorithms and exploring potentially under-utilized ones. This approach not only accelerates the model selection process but also ensures robust performance by preventing premature convergence to suboptimal solutions, making it particularly valuable for researchers working with limited computational resources.

The remainder of this paper is systematically structured to outline the comprehensive steps carried out during this research. Section~\ref{sec:related} presents a review of related work, focusing on the algorithms integral to this study. Section~\ref{sec:dataset} outlines the process of dataset collection, detailing the methodologies employed to curate the data. Section~\ref{sec:tl} elaborates on the development of the initial model, including the application of transfer learning techniques. Section~\ref{sec:approach} introduces the policy-guided algorithm selection, which forms the core of our proposed approach. Section~\ref{sec:results} details the experimental procedures and results, while section~\ref{sec:conclusion} offers concluding remarks, summarizing key findings and outlining potential avenues for future research.

\section{Related Research} \label{sec:related}
The intersection of RL and object detection has garnered increasing attention as researchers seek to optimize model selection and reduce computational overhead. Traditional object detection pipelines typically involve exhaustive evaluation of multiple pre-trained models across entire datasets, a process that is both time-consuming and resource-intensive. Recent computer vision approaches have advanced the state of object detection by balancing speed and accuracy; however, the challenge of selecting the most appropriate model for a given task remains largely unexplored.

Studies have explored the use of deep reinforcement learning (DRL) to enhance object detection by refining the region proposal process. One such approach~\cite{alpdemir2024reinforcement} proposes a cost-effective search strategy that utilizes RL to optimize navigation paths for efficient object detection. This approach focuses on minimizing search complexity while enhancing detection accuracy by guiding the model through the most relevant regions of an image or environment, ultimately improving detection performance with reduced computational overhead.

Recent works have attempted to address model selection using transfer learning and ensemble methods~\cite{sannasi2024deep}. A weighted ensemble method with transfer learning~\cite{meswal2024weighted} is a popular approach in object detection. However, the effective selection of a pre-trained model was not the focal point of these studies. Ensemble approaches~\cite{de2024hybrid} aggregate predictions from multiple models to enhance detection robustness, but often at the cost of increased inference time. While effective, these methods lack adaptive selection mechanisms capable of dynamically prioritizing high-performing models without redundant computations.
 
The integration of reinforcement learning into model selection introduces a novel paradigm. Works such as \citean{heuillet2024efficient} and \citean{benmeziane2024medical} applied neural architecture search (NAS) to automate model architecture design, demonstrating the efficacy of RL in optimizing complex neural networks. Similarly, \citean{nguyen2024laser} employed multi-armed bandit algorithms to guide model selection in NLP tasks, highlighting the potential of reward-based learning for efficient selection strategies. Our work builds upon these foundations by incorporating Upper Confidence Bound (UCB) algorithms into the object detection pipeline, facilitating adaptive pre-trained model selection. Furthermore, detection transformers, including RT-DETR~\cite{lv2023detrs}, have introduced end-to-end object detection frameworks that eliminate the need for post-processing techniques such as NMS. These architectures improve detection accuracy while maintaining computational efficiency, making them suitable candidates for integration with RL-based selection methods. Our study extends this line of research by evaluating the performance of RTDETRx, a real-time transformer-based model, within an RL-guided selection framework, demonstrating enhanced F1-scores and computational efficiency compared to baseline approaches.

\section{Animal Subjects and Biometric Collection} \label{sec:dataset}

In this section, we provide a detailed explanation of the procedures involving animal subjects and the collection of biometric data as integral components of this research project.

\subsection{Sheep Population}
On April 15, 2024, 141 Katahdin sheep ranging from 6 months to 10 years of age were placed in a Combi Clamp working system, weighed using Tru test XR5000 digital scale (Cattlesoft, Inc, College Station, TX), and spraypainted using quick shot spray for livestock marking. Sheep were then placed into an open paddock (3.5 ha) and allowed to freely graze until remote sensing imaging.

\subsection{Cattle Population}
The cattle herd comprised 68 Angus Cross animals: 41 brood cows with nursing calves, 14 replacement heifers, 3 breeding bulls, and 10 young calves. Animals were placed in a Filson Hydraulic Squeeze chute and weighed using a Tru-test XR5000 digital scale. Post-data collection, the herd was placed in a 0.25 ha paddock for subsequent imaging.






\subsection{Remote Sensing Data Acquisition \& Processing}

Remote sensing data collection utilized three primary systems: Skydio 2+ UAV (Skydio, San Mateo, CA) for Red, Green, Blue (RGB) imagery and video, MicaSense RedEdge MX sensor (MicaSense, Inc., Seattle, WA) for multispectral data (capturing five discrete bands: blue (459-491 nm), green (547-573 nm), red (661-675 nm), red edge (711-723 nm), and near infrared (814-871 nm)), and ICI 9640P radiometric thermal camera (Infrared Cameras, Inc., Beaumont, TX) for ground-based thermal imaging in (C).

Sheep imagery was collected on April 17, 2024, consisting of Red, Green, Blue (RGB) imagery and photogrammetry exclusively. The comprehensive cattle data collection occurred on May 21, 2024, incorporating RGB, multispectral, and thermal imaging. Aerial data was captured at a consistent altitude of 30 meters above ground level. RGB and multispectral imagery processing was conducted using Agisoft Metashape (Metashape, St. Petersburg, Russia) to generate 3D structure from motion (SFM) models~\cite{over2021processing}. 

In this research, we analyzed RGB images of cattle collected across 4 scanning runs, comprising 30, 31, 38, and 38 images per run, for a total of 137 images.

\subsection{Image Annotation}

For image annotation, we utilized the Computer Vision Annotation Tool (CVAT)~\cite{CVAT_ai_Corporation_Computer_Vision_Annotation_2023} to manually label the RGB images of both sheep and cattle. Each animal was annotated with a bounding box to define its spatial location and extent within the image. The annotation process involved carefully delineating rectangular regions around individual animals, with particular attention to capturing complete animal profiles, while minimizing background inclusion. The resulting annotations serve as ground truth data for evaluating the performance of our animal detection models and were exported in COCO (Common Objects in Context)~\cite{cocodataset} format, which includes the spatial coordinates of each bounding box along with corresponding class labels for sheep and cattle respectively.

\begin{figure}[t]
\centering
\includegraphics[width=0.9\columnwidth]{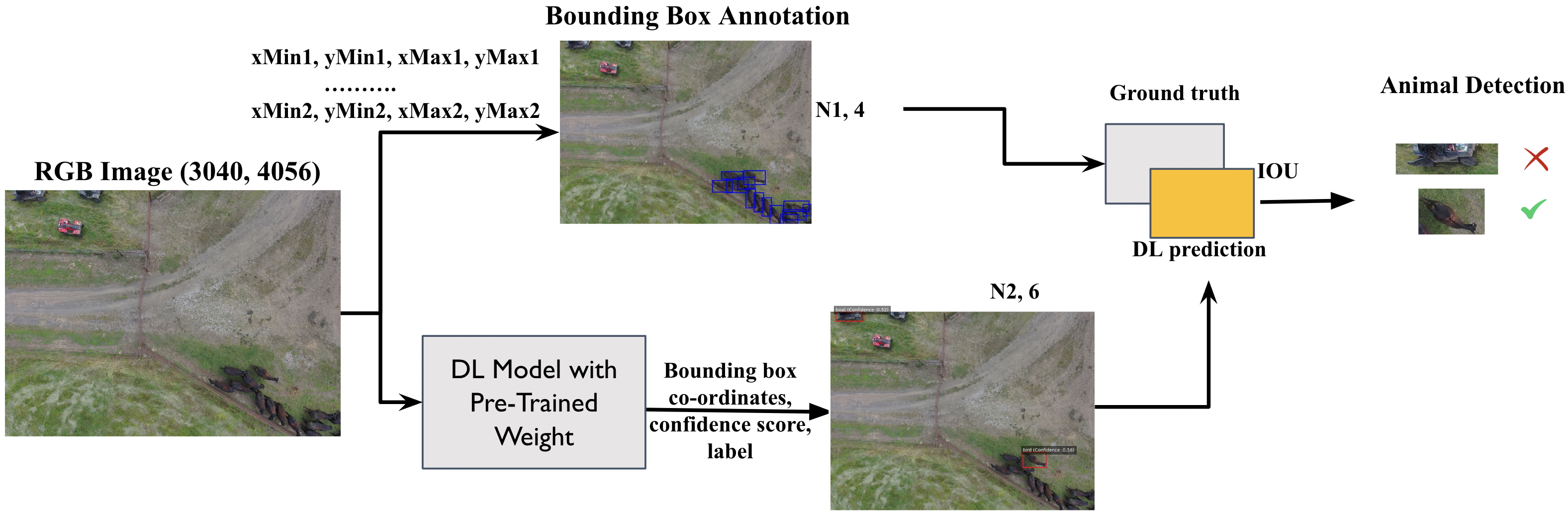}
\caption{Overview of the transfer learning approach. We create ground truth data by manually annotating object bounding boxes using CVAT, then leverage pre-trained deep learning models to predict objects in images. The models' predicted bounding boxes are compared against our ground truth annotations to evaluate the effectiveness of this approach.}
\label{fig:transfer_learning}
\end{figure}


\section{Transfer Learning} \label{sec:tl}

Transfer learning (Figure~\ref{fig:transfer_learning}) is a popular machine learning technique where a model developed for one task is reused for a model on a second task. In this study, we employed 16 established computer vision models specialized for detection and segmentation tasks, each initialized with pre-trained weights from the MS COCO~\cite{cocodataset} dataset.

\newenvironment{algocolor}{%
   \setlength{\parindent}{0pt}
   \itshape
   \color{gray}
}{}

\begin{algorithm}[bt!]
\DontPrintSemicolon
\SetAlgoLined
\SetNoFillComment
\setstretch{1.25}

\KwDefinition{ transferLearning (pre-trained model $\M_i$, animal dataset RGB image $\I_j$, overlapping threshold $\rho$, confidence threshold $\tau$, ANIMAL\_CLASSES)}

\KwResult{Ground truth, predicted, and matched bounding box coordinates for animals in the image. } \newline
\begin{algocolor}
$\B_j \gets \varnothing$\\

$\G_j \gets $ GroundTruth\_Annotation($\I_j$) \\
model $\gets$ LoadModelWithPre-trainedWeights($\M_i$)\\

$\D_j \gets $ model.predict($\M_i$, $\I_j$)\\ 

$\Z_j \gets $ BoundingBox.match($\D_j, \G_j, \rho$)\\

\For{$(b, c, s)\ in\ \Z_j$}
{ 
    \If {$s \geq \tau$ AND $c \in $ ANIMAL\_CLASSES} {
    $\B_j \gets b$\\
    }
    
}
return $\G_j, \D_j, \B_j$ \newline
\end{algocolor}
\caption{Transfer Learning using pre-trained COCO dataset weights.}\label{alg:tlearn}

\end{algorithm} 

\subsection{Problem Formulation}
We are given a finite set of RGB images representing herds of sheep and cattle. Let the total number of images be $N$. Each image is denoted as $I_i$, where $i \in {1, 2, .... N}$. Formally, the image set can be expressed as: $I = {I_1, I_2,...I_N}$. Each $I_i$ is a three-channel RGB image: $I_i : \Omega \rightarrow \mathbb{R}^3$.\newline

\noindent \textit{Ground Truth Annotations:} Each image $I_i$ contains $G_i$ ground truth animal instances, where $g_i$ is the total number of animals in $I_i$. The ground truth annotations of an image $I_i$:
\[G_i = \{(b_j^*, c_j^*) | j \in {1, 2, ... g_i} \}\]

$b_j^* \rightarrow$ the ground truth bounding box for the $j$-th animal.

$c_j^* \in \{ cow, sheep, cat, dog, horse, person, elephant, bear,$\newline $zebra, giraffe, person\} $ is the corresponding class label derived to match with animal classes from COCO dataset.  \newline 

\noindent \textit{Transfer Learning:} We apply transfer learning using pre-trained model $M_k$ over image dataset $I$ for object detection (Algorithm~\ref{alg:tlearn}). Formally, the pre-trained model set can be expressed as: $M = M_1, M_2,....M_{16}$. For each image $I_i$, the model $M_k$ generates predictions:

\[D_i = \{(b_j, c_j, s_j) | j \in {1, 2, ... M_i} \}\]

$b_j \rightarrow$ the predicted bounding box for object $j$.\newline
\indent $c_j \rightarrow$ the predicted object class label.\newline
\indent $s_j \rightarrow$ the confidence score.\newline
\indent $M_i \rightarrow$ the number of detected objects in $I_i$.\newline

\begin{figure}[t]
\centering
\includegraphics[width=0.99\columnwidth]{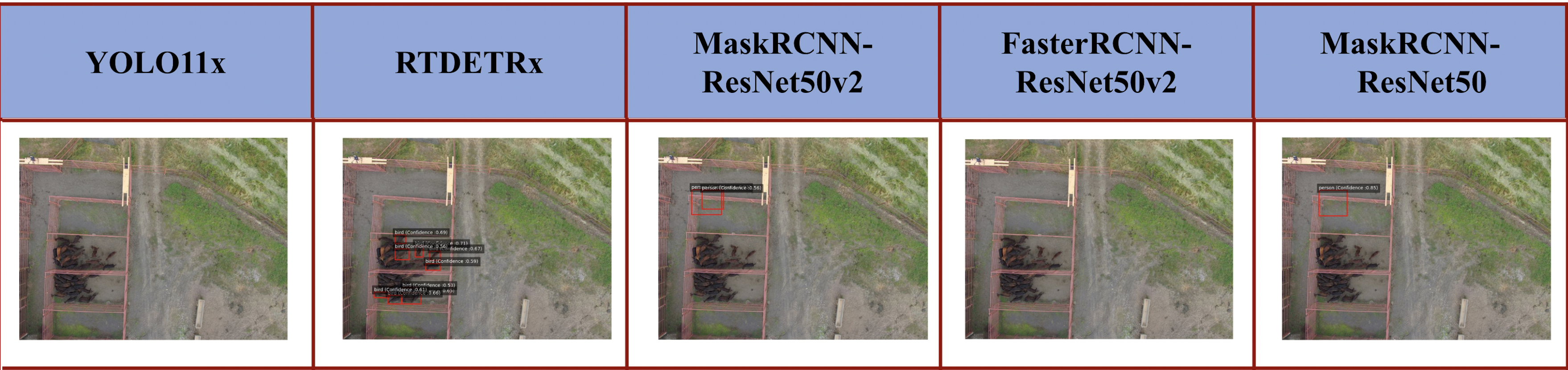}
\caption{Examples of object detection results on RGB images. Each detected object is highlighted with a red bounding box, accompanied by its class label and confidence score displayed above. Some images show no detections, indicating that certain models did not identify any objects of interest in those cases.}
\label{fig:predictions}
\end{figure}

\noindent \textit{Animal Detection:}
Pre-trained model performance on animal detection is evaluated based on the following conditions: 

\begin{itemize}
    \item The Intersection over Union (IoU) between the predicted bounding box $b_j$ and ground truth $b_j^*$ must exceed the overlapping threshold, $\rho$.
    \item The confidence score, $s_j$ must be greater than $\tau$.
    \item The predicted class $c_j$ must belong to the set of animal classes defined for $c_j^*$ 
\end{itemize}

The objective is to identify the optimal model that maximizes the F1 score, ensuring precise detection and segmentation of animal herds despite the constraints of limited training data. The focus is on minimizing false positives while maintaining high recall to accurately detect all annotated instances.

\subsection{Pre-trained Models} \label{subsec:pre-trained}

We employed a diverse selection of detection and segmentation models to generate bounding boxes and object detection scores for our animal RGB image data. High-precision segmentation models, fast detection models, and lightweight architectures were chosen to ensure an optimal balance between accuracy, speed, and computational efficiency. All pretrained models were sourced from PyTorch Vision~\cite{paszke2017automatic} and Ultralytics~\cite{ultralytics}.

\begin{figure}[t]
\centering
\includegraphics[width=0.99\columnwidth]{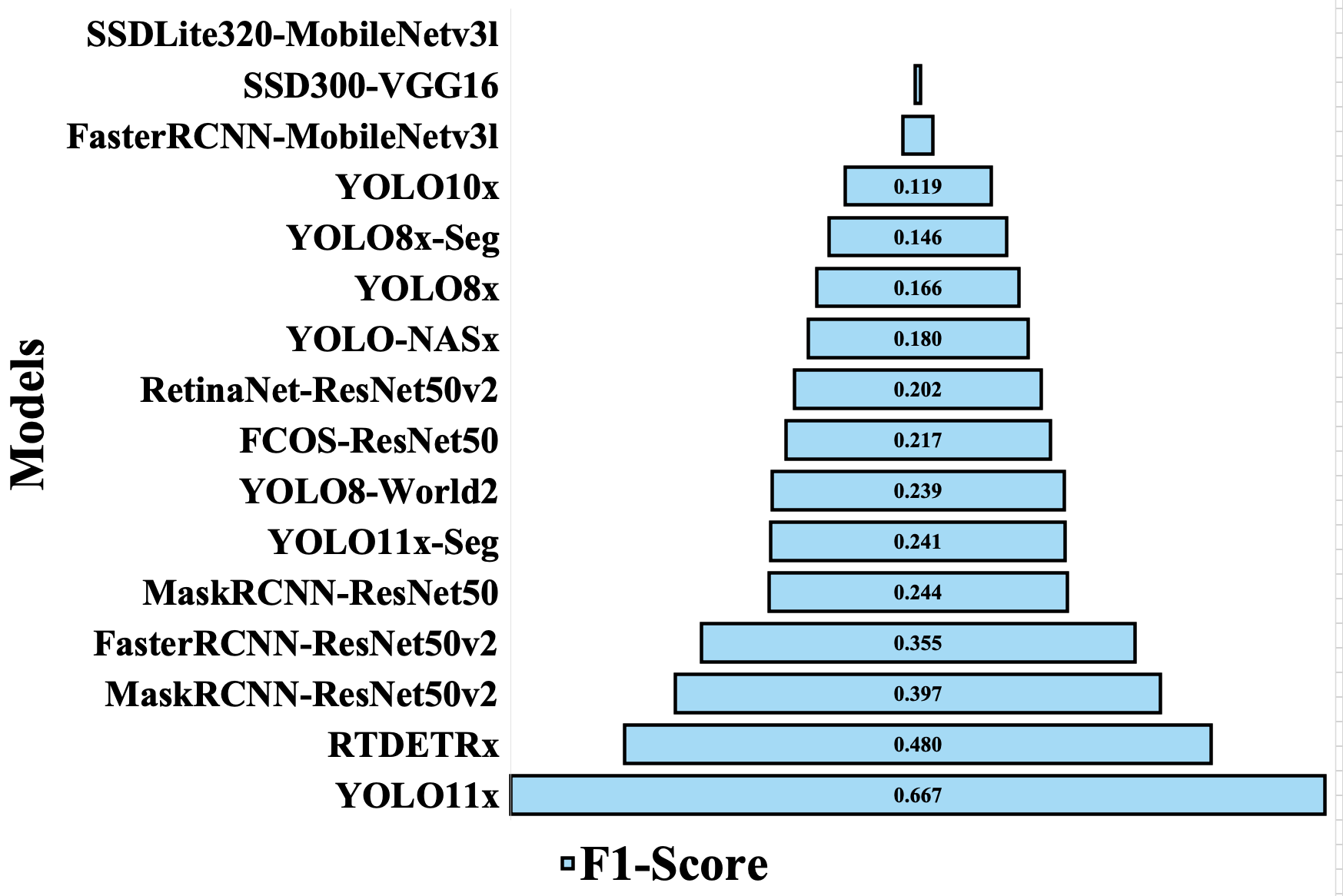}
\caption{Using pre-trained weights from the COCO dataset, we evaluated 16 different deep learning models on our image collection. Model names are specified in section~\ref{subsec:pre-trained}. }
\label{fig:transfer_learning_evaluation}
\end{figure}

\subsubsection{YOLO Models (You Only Look Once)} \label{yolo}
YOLO~\cite{redmon2016you} models are renowned for their real-time object detection capabilities, which strike a balance between accuracy and speed~\cite{ali2024advancing}. They predict bounding boxes and class probabilities directly from full images, enabling fast inference. YOLO8x-Seg~\cite{varghese2024yolov8} is a segmentation variant that performs instance segmentation, identifying object masks in addition to bounding boxes. YOLO8x is a standard detection model that predicts bounding boxes for objects without segmentation. YOLO10x and YOLO11x~\cite{wang2024yolov10} represent experimental or customized iterations integrating larger architectures or improved training datasets to boost accuracy further. YOLO11x-Seg combines segmentation capabilities with the enhancements from the 11x iteration. YOLO-NASx~\cite{supergradients} integrates neural architecture search (NAS) principles with the YOLO object detection framework, automatically discovering optimal network architectures for detection tasks. YOLO8-World2~\cite{cheng2024yolow} refers to YOLOWorld, a version optimized for open-vocabulary detection, capable of detecting objects from categories not explicitly present in the training dataset. 

\subsubsection{RT-DETR (Real-Time DEtection TRansformer)}

RT-DETR~\cite{lv2023detrs} introduces transformer-based detection models that are faster and more accurate compared to traditional architectures. It integrates a hybrid encoder-decoder structure, combining convolutional neural networks (CNNs) for feature extraction with transformer-based attention mechanisms to refine object localization and classification.

\subsubsection{Mask R-CNN (Region-Based Convolutional Neural Network)}
Mask R-CNN~\cite{he2017mask} performs instance segmentation, predicting object masks along with bounding boxes. MaskRCNN-ResNet50 and MaskRCNN-ResNet50v2 use ResNet-50~\cite{he2016deep} as the backbone feature extractor. The ``v2'' denotes an improved version with enhanced weight initialization, faster convergence, and higher accuracy. Mask R-CNN is widely used in medical imaging~\cite{chen2021liver}, autonomous driving~\cite{ojha2021vehicle}, and scenarios requiring precise object boundaries. Faster R-CNN (Fast Region-Based Convolutional Network)~\cite{ren2015faster} is designed for object detection without segmentation, offering high accuracy by utilizing a region proposal network (RPN). It detects bounding boxes but not object masks. FasterRCNN-ResNet50v2 incorporates ResNet-50 (v2) as the backbone, providing state-of-the-art accuracy with optimized speed. FasterRCNN-MobileNetv3l is a lightweight Faster R-CNN variant with MobileNetv3 Large~\cite{howard2017mobilenets}, balancing speed and performance, ideal for deployment on mobile or edge devices.

\subsubsection{FCOS (Fully Convolutional One-Stage Object Detection)}
FCOS-ResNet50~\cite{tian2019fcos} is an anchor-free object detection model that simplifies the detection pipeline by directly predicting object locations, eliminating the need for predefined anchor boxes. It performs well on dense object detection tasks.

\subsubsection{RetinaNet}
RetinaNet-ResNet50v2~\cite{li2019light} uses a ResNet-50 backbone and introduces Focal Loss to handle class imbalance, making it effective in detecting small or rare objects in large datasets.

\subsubsection{SSD (Single Shot MultiBox Detector)}
SSD~\cite{liu2016ssd} performs object detection in a single forward pass, making it faster than two-stage methods like Faster R-CNN. SSD300-VGG16 -
SSD300, with a VGG-16~\cite{simonyan2014very} backbone, is optimized for speed and accuracy in detecting medium to large objects. SSDLite320-MobileNetv3l is a lightweight SSD variant using MobileNetv3 Large and optimized for mobile/embedded systems, achieving fast inference with reduced computational cost.

\subsection{Preliminary Experiments}
To select the model that achieves the highest F1 score, we performed multiple consensus-based evaluations across the dataset~(Table~\ref{tab:tl_evaluations}). This iterative process ensured robust assessment and comparison, leading to the identification of the most effective model for detection and segmentation tasks. We employed the complete set of images to conduct a comprehensive evaluation of these baselines.

\begin{table}[t!]
 \centering
 \scriptsize
 \caption{Transfer learning experiments with varying overlapping threshold, confidence score, and overlapping boxes.}
  \begin{tabular}{>{\columncolor[gray]{0.9}}l|c|c|c}
    \rowcolor{cyan!20}
 \mystrut     Criteria & Precision & Recall & F1-Score\\ \hline
      
    \midrule
$\rho \geq $ 0.5 \& $\tau \geq $ 0.5 & 0.817 & 0.634 & 0.673 \\ \hline
$\rho \geq $ 0.6 \& $\tau \geq $ 0.5 & 0.838 & 0.559 & 0.625 \\ \hline
$\tau \geq$ 0.01 &  0.87 & 0.51 & 0.59 \\ \hline
\makecell[l]{Atleast 3 bounding box predictions\\overlap with $\rho \geq $ 0.5} & 0.826 & 0.63 & 0.674 \\
    \bottomrule
   \end{tabular}
 
\label{tab:tl_evaluations}
\end{table}
\begin{itemize}
    \item As a baseline, model evaluation was conducted based on three criteria: Intersection over Union (IoU) greater than 50\%, the predicted class label belonging to animal classes, and a confidence score exceeding 50\%. Figure~\ref{fig:transfer_learning_evaluation} presents the overall F1 score comparison across all 16 models, with YOLO11x emerging as the highest-performing model. Figure~\ref{fig:predictions} shows some of the prediction samples.

    \item As a slight modification to the evaluation metrics, we experimented with varying confidence score thresholds. When evaluated with $\tau \geq$ 0.6, the model's performance decreased to 0.625.

    \item Another evaluation was conducted by reducing the IoU overlap threshold $\rho$ to 0.01. While this adjustment led to an increase in precision, recall declined, resulting in a decrease in the F1 score to 0.51 and 0.59, respectively.

    \item In the next evaluation, predictions from all models were fused. A bounding box was selected only if it overlapped with at least three model predictions, with an IoU threshold of 0.5. Additionally, the confidence score for the object being classified as an animal was required to be at least 0.5. This approach did not yield significant improvements compared to the previous model.
    
\end{itemize}
 
All these evaluations necessitate running every model on the entire set of images, which involves extensive computational resources and time. The process of selecting a single model that yields the highest performance requires exhaustive comparisons, making it a labor-intensive and time-consuming task. Each iteration involves assessing multiple models against various metrics, such as IoU thresholds, confidence scores, and class predictions, further extending the evaluation period.

This highlights the critical need for a more efficient learning framework capable of reducing the number of iterations while still identifying the most suitable model for animal detection tasks. By minimizing redundant evaluations, such a framework could expedite deployment in real-world applications, ensuring rapid detection of animal herd populations and health, even when working with limited data or constrained computational environments.

\begin{figure}[t]
\centering
\includegraphics[width=0.95\columnwidth, trim={0 0 0 0},clip]{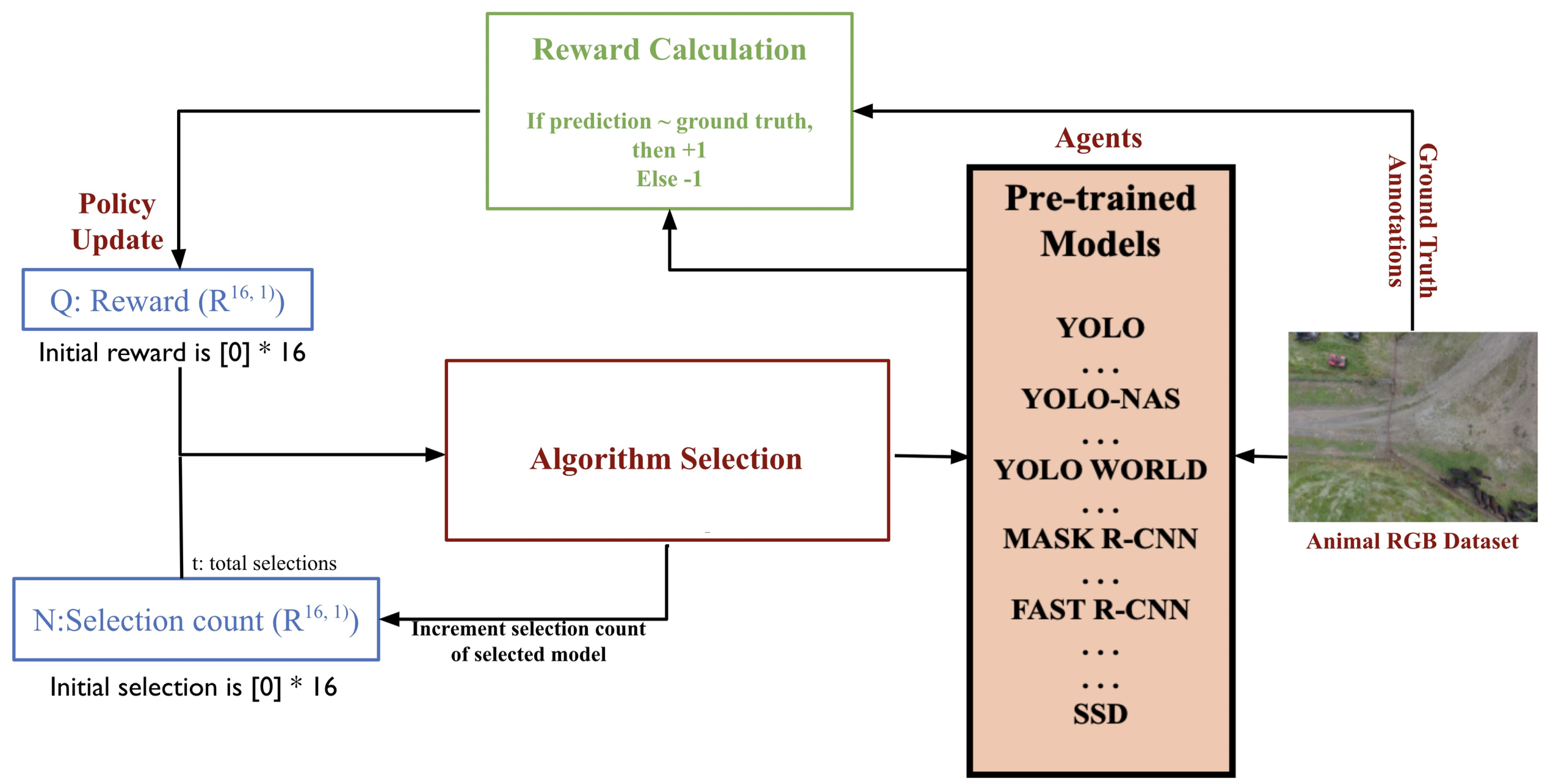}
\caption{Overview of the Reinforcement learning (RL) Process}
\label{fig:rl_learning}
\end{figure}

\begin{algorithm}[t]
\DontPrintSemicolon
\SetAlgoLined
\SetNoFillComment
\setstretch{1.2}

\KwDefinition{ algorithmSelector (totalNumberOfModels $n$, totalSteps $t$, rewardVector $\Q$, selectionVector $\N$, explorationCostant $\C$)}

\KwResult{Model index selected by UCB approach} 
\begin{algocolor}

$\UCB \gets \varnothing$\\

\For{$action\ a\ in\ 1..n$}
{
    \If{selectionVector[$a$] is never selected} {
        return $a$\\
    } {
    $\UCB_a \gets \Q_a + \C * \sqrt{ln(t)/\N_a}$ \\
    }
}
return $\underset{a}{\argmax}\ \UCB$ \newline
\end{algocolor}
\caption{Upper Confidence Bound (UCB) based model selection}\label{alg:ucb_selection}
\end{algorithm}

\begin{algorithm}[t!]
\DontPrintSemicolon
\SetAlgoLined
\SetNoFillComment
\setstretch{1.2}

\KwDefinition{ reinforcementBasedAnimalDetection (models $\M$, animal dataset, $\I$, 
ANIMAL\_CLASSES)}

\KwResult{Index of best-performing Pre-trained Model} 
\begin{algocolor}

$n \gets $Number of models, $\M$\\
$\Q \gets [0]$ * n \Comment{reward vector}
$\N \gets [0]$ * n \Comment{selection vector}
$\C \gets $ 0.1 \Comment{exploration constant}

$\rho \gets $ 0.5 \Comment{overlapping threshold}
$\tau \gets $ 0.5 \Comment{detection confidence threshold}

\For{$image\ I_j\ in\ \I$}
{

$t \gets  \sum_{i=1}^{n} \N_{i} $ \Comment{total number of selections}
$a \gets $algorithmSelector($n$, $t$, $\Q$, $\N$, $\C$)\newline
\Comment{index of selected model}
$\N_a += 1$ \Comment{increment selection count}
$\G_j, \D_j, \B_j \gets $transferLearning ( $\M[a]$, $\I_j$, $\G_j$, $\rho$, $\tau$, ANIMAL\_CLASSES)\\
$b \gets $ size of $\B_j$\\
$g \gets $ size of $\G_j$\\
$d \gets $ size of $\D_j$\\
$\Q_a += b$ \Comment{rewards for correct selections}
$\Q_a -=  g - b$ \Comment{penalty for false negatives}
$\Q_a -=  d - b$ \Comment{penalty for false positives}
}
\Comment{Return highest reward model index}
return $\underset{a}{\argmax}\ \Q$ \newline
\end{algocolor}
\caption{Reinforcement Learning based animal detection model selection }\label{alg:policy_guided_animalDetection}
\end{algorithm}

\section{Policy Guided Model Selection} \label{sec:approach}

To efficiently address the challenge of identifying the most effective pre-trained model without conducting exhaustive evaluations across all models and images, we leverage reinforcement learning (RL) driven by the Upper Confidence Bound (UCB) algorithm (Figure~\ref{fig:rl_learning}).  Reinforcement learning is a branch of machine learning where agents learn to make decisions by interacting with an environment and receiving feedback through rewards or penalties. This iterative process enables the agent to optimize its actions over time, maximizing cumulative rewards. Traditional methods for selecting the best-performing model typically requires running each pre-trained model over the entire dataset, resulting in excessive computational costs and time. This brute-force approach, while thorough, is impractical for large-scale datasets or environments where real-time decision-making is essential. Our proposed method introduces a reward-penalty framework that dynamically balances exploration and exploitation, allowing the system to converge on the optimal model without unnecessary computations. By using RL principles, we ensure that model selection iterates intelligently, significantly reducing the overall number of model executions.

The process (Algorithm~\ref{alg:policy_guided_animalDetection}) begins with the initialization of two key vectors: the selection vector and the reward vector. The selection vector acts as a counter, recording how often each model is chosen for evaluation, while the reward vector tracks cumulative rewards earned by each model over the course of the evaluation. Both vectors are initialized to zero to ensure that no model has an advantage at the outset. The UCB algorithm (Algorithm~\ref{alg:ucb_selection}) played a critical role in model selection by computing confidence scores that guide the decision-making process. The algorithm prioritized actions based on two factors: the average performance of each model (exploitation) and the degree of uncertainty or lack of exploration (exploration). An exploration constant dictates the balance between trying new models and refining the performance of models that already perform well.

Initially, UCB ensures that each model is executed at least once to establish baseline performance metrics. If a model has not been selected, it receives priority until all models have been evaluated. Once this initial phase is complete, the model with the highest UCB score was selected for further evaluation. Upon selection, the chosen model is executed to perform object detection on an image, producing bounding boxes, confidence scores, and class predictions. This step mirrors standard transfer learning, where the pre-trained model was fine-tuned for the specific task at hand. After the execution, the model's performance was assessed by comparing its predictions to the ground truth labels. Rewards were assigned for correctly matched bounding boxes, while penalties are applied for false negatives (missed objects), false positives (incorrect detections), and misclassification. Following each evaluation, the selection count for the executed model is incremented, and the reward vector was updated.

By the end of the process, the model with the highest cumulative reward was identified as the best-performing model. This model can then be deployed directly for further use or undergo additional fine-tuning to improve performance further. The advantage of this approach lies in its efficiency – the RL-UCB framework not only minimizes redundant computations but also adapts dynamically, ensuring that promising models are prioritized without neglecting underexplored ones.

\section{Experiments \& Results} \label{sec:results}

\begin{table}[b!]
 \centering
 \scriptsize
 \caption{Reinforcement learning based evaluations}
  \begin{tabular}{>{\columncolor[gray]{0.9}}l|c|c|c}
    \rowcolor{cyan!20}
 \mystrut     Criteria & Precision & Recall & F1-Score\\ \hline
      
    \midrule
 Brute-force RL - one model & 0.82825 & 0.673 & 0.69 \\ \hline
Brute-force RL - Consensus from 2 models & 0.709 & 0.479 & 0.517 \\ \hline

\textbf{Our Method - RL with UCB} & 0.795 & 0.714 & \textbf{0.718}\\
    \bottomrule
   \end{tabular}
 
\label{tab:rl_evaluations}
\end{table}

During transfer learning with 16 models utilizing pre-trained weights from the COCO dataset, Figure~\ref{fig:transfer_learning_evaluation} illustrates the F1-score performance metric achieved by iteratively evaluating all methods across the entire dataset. The results identify the top-performing classifiers as YOLOv11x and RT-DETR. RT-DETR is engineered to enhance the inference speed of transformer-based models while maintaining high accuracy. Conversely, YOLO is highly optimized for rapid convergence during training, often delivering robust results with reduced computational costs and smaller datasets compared to transformer-based architectures. The findings suggest that both models are advantageous, with their utility dependent on the specific requirements of the application.

To minimize redundant computations while ensuring the selection of promising models through an effective search process, we implemented our proposed reinforcement learning (RL) framework and conducted a series of experiments. Specifically, we performed three distinct RL-based experiments to evaluate the effectiveness of our search method in identifying the best-performing pre-trained model for object detection~(Table~\ref{tab:rl_evaluations}). The entire dataset was partitioned into training and test sets using a 90:10 ratio. Each evaluation was conducted four times, and the results were averaged to obtain a comprehensive assessment score.

The first approach employed a brute-force RL method, bypassing algorithmic selection techniques such as Upper Confidence Bound (UCB) or random selection. Instead, all models were executed on the entire dataset, and rewards were computed post hoc. This exhaustive method selected RTDETRx as the optimal model, achieving high precision but yielding average recall. While this approach provided a useful baseline, it offered no computational efficiency improvements over traditional exhaustive search methods.

The second approach adopted a consensus-based strategy, wherein bounding boxes were generated by all pre-trained models, and those with agreement from at least two models were selected for reward calculation. Despite the novelty of this approach, it resulted in the lowest recall and subsequently the weakest F1-score, underscoring its limitations in high-recall tasks.

Finally, we implemented our proposed method, integrating reinforcement learning with UCB-based pre-trained model selection. By executing a single model per image rather than applying brute-force evaluation across all models, this approach significantly reduced computational overhead while achieving the highest F1-score among the three methods. Notably, RTDETRx emerged as the top-performing model across all approaches, demonstrating superior performance in detecting animals from limited RGB data. 

This outcome highlights the effectiveness of transfer learning using pre-trained weights from the COCO dataset, reinforcing the potential of combining pre-trained computer vision algorithms with reinforcement learning for efficient object detection in resource-constrained environments.

\section{Conclusion} \label{sec:conclusion}

In this study, we explored the integration of RL with pre-trained model selection to optimize object detection tasks, addressing the computational inefficiencies inherent in exhaustive model evaluation. Through several baseline approaches, we demonstrated the impact of algorithmic selection on performance metrics such as precision, recall, and F1-score. Our final method, which leveraged Upper Confidence Bound (UCB)-based RL, significantly reduced the number of model executions while achieving superior detection accuracy. This approach not only streamlined the model selection process but also highlighted the effectiveness of RTDETRx in detecting objects from limited RGB data, thereby reinforcing the potential of transformer-based architectures in real-time applications. The results underscore the value of balancing exploration and exploitation in model selection, paving the way for more efficient deployment of pre-trained models in resource-constrained environments. Future work will focus on extending this framework to multi-task learning scenarios and incorporating domain-specific datasets to further enhance generalization and adaptability. Future work necessitates the augmentation of large-scale data from limited resources, alongside the development of task-specific neural network models tailored for animal population and health detection. These models can subsequently be employed for real-time and efficient animal monitoring applications.

\section*{Acknowledgment}

This research was supported by the USDA-ARS agreement ``Developing Detection and Modeling Tools for the Geospatial and Environmental Epidemiology of Animal Disease'' \#58-6064-3-017. The USDA is an equal opportunity provider and employer. Mention of trade names or commercial products in this article is solely for the purpose of providing specific information and does not imply recommendation or endorsement by the USDA. Animal data collection was made possible through the assistance of Erin Wood, Brent Wooley, Chad Lee, Jacob Nickols, and Michael Schmidt with the USDA-ARS and Malcom Williamson with the University of Arkansas. 

\bibliography{reference}

\begin{thebibliography}{40}
\providecommand{\natexlab}[1]{#1}
\providecommand{\url}[1]{#1}
\csname url@samestyle\endcsname
\providecommand{\newblock}{\relax}
\providecommand{\bibinfo}[2]{#2}
\providecommand{\BIBentrySTDinterwordspacing}{\spaceskip=0pt\relax}
\providecommand{\BIBentryALTinterwordstretchfactor}{4}
\providecommand{\BIBentryALTinterwordspacing}{\spaceskip=\fontdimen2\font plus
\BIBentryALTinterwordstretchfactor\fontdimen3\font minus \fontdimen4\font\relax}
\providecommand{\BIBforeignlanguage}[2]{{%
\expandafter\ifx\csname l@#1\endcsname\relax
\typeout{** WARNING: IEEEtranN.bst: No hyphenation pattern has been}%
\typeout{** loaded for the language `#1'. Using the pattern for}%
\typeout{** the default language instead.}%
\else
\language=\csname l@#1\endcsname
\fi
#2}}
\providecommand{\BIBdecl}{\relax}
\BIBdecl

\bibitem[Norouzzadeh et~al.(2018)Norouzzadeh, Nguyen, Kosmala, Swanson, Palmer, Packer, and Clune]{norouzzadeh2018automatically}
M.~S. Norouzzadeh, A.~Nguyen, M.~Kosmala, A.~Swanson, M.~S. Palmer, C.~Packer, and J.~Clune, ``Automatically identifying, counting, and describing wild animals in camera-trap images with deep learning,'' \emph{Proceedings of the National Academy of Sciences}, 2018.

\bibitem[Green et~al.(2005)Green, Hastings, Arzberger, Ayala, Cottingham, Cuddington, Davis, Dunne, Fortin, Gerber, et~al.]{green2005complexity}
J.~L. Green, A.~Hastings, P.~Arzberger, F.~J. Ayala, K.~L. Cottingham, K.~Cuddington, F.~Davis, J.~A. Dunne, M.-J. Fortin, L.~Gerber \emph{et~al.}, ``Complexity in ecology and conservation: mathematical, statistical, and computational challenges,'' \emph{BioScience}, 2005.

\bibitem[K{\"u}hl(2008)]{kuhl2008best}
H.~K{\"u}hl, \emph{Best practice guidelines for the surveys and monitoring of great ape populations}.\hskip 1em plus 0.5em minus 0.4em\relax IUCN, 2008.

\bibitem[Buchelt et~al.(2024)Buchelt, Adrowitzer, Kieseberg, Gollob, Nothdurft, Eresheim, Tschiatschek, Stampfer, and Holzinger]{buchelt2024exploring}
A.~Buchelt, A.~Adrowitzer, P.~Kieseberg, C.~Gollob, A.~Nothdurft, S.~Eresheim, S.~Tschiatschek, K.~Stampfer, and A.~Holzinger, ``Exploring artificial intelligence for applications of drones in forest ecology and management,'' \emph{Forest Ecology and Management}, 2024.

\bibitem[Kumar et~al.(2023)Kumar, Luo, and Shaukat]{kumar2023comprehensive}
P.~Kumar, S.~Luo, and K.~Shaukat, ``A comprehensive review of deep learning approaches for animal detection on video data.'' \emph{International Journal of Advanced Computer Science \& Applications}, vol.~14, no.~11, 2023.

\bibitem[Chen et~al.(2023)Chen, Jacob, Shoshani, and Charter]{chen2023using}
A.~Chen, M.~Jacob, G.~Shoshani, and M.~Charter, ``Using computer vision, image analysis and uavs for the automatic recognition and counting of common cranes (grus grus),'' \emph{Journal of Environmental Management}, 2023.

\bibitem[Petso et~al.(2022)Petso, Jamisola~Jr, and Mpoeleng]{petso2022review}
T.~Petso, R.~S. Jamisola~Jr, and D.~Mpoeleng, ``Review on methods used for wildlife species and individual identification,'' \emph{European Journal of Wildlife Research}, vol.~68, no.~1, p.~3, 2022.

\bibitem[Besson et~al.(2022)Besson, Alison, Bjerge, Gorochowski, H{\o}ye, Jucker, Mann, and Clements]{besson2022towards}
M.~Besson, J.~Alison, K.~Bjerge, T.~E. Gorochowski, T.~T. H{\o}ye, T.~Jucker, H.~M. Mann, and C.~F. Clements, ``Towards the fully automated monitoring of ecological communities,'' \emph{Ecology Letters}, 2022.

\bibitem[Hosain et~al.(2024)Hosain, Zaman, Abir, Akter, Mursalin, and Khan]{hosain2024synchronizing}
M.~T. Hosain, A.~Zaman, M.~R. Abir, S.~Akter, S.~Mursalin, and S.~S. Khan, ``Synchronizing object detection: Applications, advancements and existing challenges,'' \emph{IEEE Access}, 2024.

\bibitem[Pal et~al.(2021)Pal, Pramanik, Maiti, and Mitra]{pal2021deep}
S.~K. Pal, A.~Pramanik, J.~Maiti, and P.~Mitra, ``Deep learning in multi-object detection and tracking: state of the art,'' \emph{Applied Intelligence}, 2021.

\bibitem[Khanam et~al.(2024)Khanam, Hussain, Hill, and Allen]{khanam2024comprehensive}
R.~Khanam, M.~Hussain, R.~Hill, and P.~Allen, ``A comprehensive review of convolutional neural networks for defect detection in industrial applications,'' \emph{IEEE Access}, 2024.

\bibitem[Alpdemir and Sezgin(2024)]{alpdemir2024reinforcement}
M.~N. Alpdemir and M.~Sezgin, ``A reinforcement learning (rl)-based hybrid method for ground penetrating radar (gpr)-driven buried object detection,'' \emph{Neural Computing and Applications}, vol.~36, no.~14, pp. 8199--8219, 2024.

\bibitem[Sannasi~Chakravarthy et~al.(2024)Sannasi~Chakravarthy, Bharanidharan, Vinoth~Kumar, Mahesh, Alqahtani, and Guluwadi]{sannasi2024deep}
S.~Sannasi~Chakravarthy, N.~Bharanidharan, V.~Vinoth~Kumar, T.~Mahesh, M.~S. Alqahtani, and S.~Guluwadi, ``Deep transfer learning with fuzzy ensemble approach for the early detection of breast cancer,'' \emph{BMC Medical Imaging}, 2024.

\bibitem[Meswal et~al.(2024)Meswal, Kumar, Gupta, and Roy]{meswal2024weighted}
H.~Meswal, D.~Kumar, A.~Gupta, and S.~Roy, ``A weighted ensemble transfer learning approach for melanoma classification from skin lesion images,'' \emph{Multimedia Tools and Applications}, 2024.

\bibitem[De~Oliveira et~al.(2024)De~Oliveira, do~Nascimento, Roberto, Tosta, Martins, and Neves]{de2024hybrid}
C.~I. De~Oliveira, M.~Z. do~Nascimento, G.~F. Roberto, T.~A. Tosta, A.~S. Martins, and L.~A. Neves, ``Hybrid models for classifying histological images: An association of deep features by transfer learning with ensemble classifier,'' \emph{Multimedia Tools and Applications}, 2024.

\bibitem[Heuillet et~al.(2024)Heuillet, Nasser, Arioui, and Tabia]{heuillet2024efficient}
A.~Heuillet, A.~Nasser, H.~Arioui, and H.~Tabia, ``Efficient automation of neural network design: A survey on differentiable neural architecture search,'' \emph{ACM Computing Surveys}, vol.~56, no.~11, pp. 1--36, 2024.

\bibitem[Benmeziane et~al.(2024)Benmeziane, Hamzaoui, Cherif, and El~Maghraoui]{benmeziane2024medical}
H.~Benmeziane, I.~Hamzaoui, Z.~Cherif, and K.~El~Maghraoui, ``Medical neural architecture search: Survey and taxonomy,'' in \emph{International Joint Conference on Artificial Intelligence}, 2024.

\bibitem[Nguyen et~al.(2024)Nguyen, Prasad, Stengel-Eskin, and Bansal]{nguyen2024laser}
D.~Nguyen, A.~Prasad, E.~Stengel-Eskin, and M.~Bansal, ``Laser: Learning to adaptively select reward models with multi-armed bandits,'' 2024.

\bibitem[Lv et~al.(2023)Lv, Xu, Zhao, Wang, Wei, Cui, Du, Dang, and Liu]{lv2023detrs}
W.~Lv, S.~Xu, Y.~Zhao, G.~Wang, J.~Wei, C.~Cui, Y.~Du, Q.~Dang, and Y.~Liu, ``Detrs beat yolos on real-time object detection,'' 2023.

\bibitem[Over et~al.(2021)Over, Ritchie, Kranenburg, Brown, Buscombe, Noble, Sherwood, Warrick, and Wernette]{over2021processing}
J.-S.~R. Over, A.~C. Ritchie, C.~J. Kranenburg, J.~A. Brown, D.~D. Buscombe, T.~Noble, C.~R. Sherwood, J.~A. Warrick, and P.~A. Wernette, ``Processing coastal imagery with agisoft metashape professional edition, version 1.6—structure from motion workflow documentation,'' US Geological Survey, Tech. Rep., 2021.

\bibitem[{CVAT.ai Corporation}(2023)]{CVAT_ai_Corporation_Computer_Vision_Annotation_2023}
{CVAT.ai Corporation}, ``{Computer Vision Annotation Tool (CVAT)},'' Nov. 2023.

\bibitem[Lin et~al.(2014)Lin, Maire, Belongie, Bourdev, Girshick, Hays, Perona, Ramanan, Doll{'{a} }r, and Zitnick]{cocodataset}
T.~Lin, M.~Maire, S.~J. Belongie, L.~D. Bourdev, R.~B. Girshick, J.~Hays, P.~Perona, D.~Ramanan, P.~Doll{'{a} }r, and C.~L. Zitnick, ``Microsoft {COCO:} common objects in context,'' \emph{CoRR}, vol. abs/1405.0312, 2014.

\bibitem[Paszke et~al.(2017)Paszke, Gross, Chintala, Chanan, Yang, DeVito, Lin, Desmaison, Antiga, and Lerer]{paszke2017automatic}
A.~Paszke, S.~Gross, S.~Chintala, G.~Chanan, E.~Yang, Z.~DeVito, Z.~Lin, A.~Desmaison, L.~Antiga, and A.~Lerer, ``Automatic differentiation in pytorch,'' 2017.

\bibitem["Glenn~Jocher(2023)]{ultralytics}
A.~C. "Glenn~Jocher, “Jing~Qiu, ``"ultralytics",'' 2023.

\bibitem[Redmon et~al.(2016)Redmon, Divvala, Girshick, and Farhadi]{redmon2016you}
J.~Redmon, S.~Divvala, R.~Girshick, and A.~Farhadi, ``You only look once: Unified, real-time object detection,'' in \emph{Proceedings of the IEEE conference on computer vision and pattern recognition}, 2016, pp. 779--788.

\bibitem[Ali et~al.(2024)Ali, Aly, Raslan, Gheith, and Amin]{ali2024advancing}
M.~A.~M. Ali, T.~Aly, A.~T. Raslan, M.~Gheith, and E.~A. Amin, ``Advancing crowd object detection: A review of yolo, cnn and vits hybrid approach,'' \emph{Journal of Intelligent Learning Systems and Applications}, 2024.

\bibitem[Varghese and Sambath(2024)]{varghese2024yolov8}
R.~Varghese and M.~Sambath, ``Yolov8: A novel object detection algorithm with enhanced performance and robustness,'' in \emph{2024 International Conference on Advances in Data Engineering and Intelligent Computing Systems (ADICS)}.\hskip 1em plus 0.5em minus 0.4em\relax IEEE, 2024, pp. 1--6.

\bibitem[Wang et~al.(2024)Wang, Chen, Liu, Chen, Lin, Han, and Ding]{wang2024yolov10}
A.~Wang, H.~Chen, L.~Liu, K.~Chen, Z.~Lin, J.~Han, and G.~Ding, ``Yolov10: Real-time end-to-end object detection,'' \emph{arXiv preprint arXiv:2405.14458}, 2024.

\bibitem[Aharon et~al.(2021)Aharon, {Louis-Dupont}, {Ofri Masad}, Yurkova, {Lotem Fridman}, {Lkdci}, Khvedchenya, Rubin, Bagrov, Tymchenko, Keren, Zhilko, and {Eran-Deci}]{supergradients}
S.~Aharon, {Louis-Dupont}, {Ofri Masad}, K.~Yurkova, {Lotem Fridman}, {Lkdci}, E.~Khvedchenya, R.~Rubin, N.~Bagrov, B.~Tymchenko, T.~Keren, A.~Zhilko, and {Eran-Deci}, ``Super-gradients,'' 2021.

\bibitem[Cheng et~al.(2024)Cheng, Song, Ge, Liu, Wang, and Shan]{cheng2024yolow}
T.~Cheng, L.~Song, Y.~Ge, W.~Liu, X.~Wang, and Y.~Shan, ``Yolo-world: Real-time open-vocabulary object detection,'' \emph{arXiv preprint arXiv:2401.17270}, 2024.

\bibitem[He et~al.(2017)He, Gkioxari, Doll{\'a}r, and Girshick]{he2017mask}
K.~He, G.~Gkioxari, P.~Doll{\'a}r, and R.~Girshick, ``Mask r-cnn,'' in \emph{Proceedings of the IEEE international conference on computer vision}, 2017, pp. 2961--2969.

\bibitem[He et~al.(2016)He, Zhang, Ren, and Sun]{he2016deep}
K.~He, X.~Zhang, S.~Ren, and J.~Sun, ``Deep residual learning for image recognition,'' in \emph{Proceedings of the IEEE conference on computer vision and pattern recognition}, 2016, pp. 770--778.

\bibitem[Chen et~al.(2021)Chen, Wei, Tang, Liu, Lai, Zhu, and He]{chen2021liver}
X.~Chen, X.~Wei, M.~Tang, A.~Liu, C.~Lai, Y.~Zhu, and W.~He, ``Liver segmentation in ct imaging with enhanced mask region-based convolutional neural networks,'' \emph{Annals of translational medicine}, vol.~9, no.~24, 2021.

\bibitem[Ojha et~al.(2021)Ojha, Sahu, and Dewangan]{ojha2021vehicle}
A.~Ojha, S.~P. Sahu, and D.~K. Dewangan, ``Vehicle detection through instance segmentation using mask r-cnn for intelligent vehicle system,'' in \emph{2021 5th international conference on intelligent computing and control systems (ICICCS)}.\hskip 1em plus 0.5em minus 0.4em\relax IEEE, 2021, pp. 954--959.

\bibitem[Ren et~al.(2015)Ren, He, Girshick, and Sun]{ren2015faster}
S.~Ren, K.~He, R.~Girshick, and J.~Sun, ``Faster r-cnn: Towards real-time object detection with region proposal networks,'' \emph{Advances in neural information processing systems}, vol.~28, 2015.

\bibitem[Howard et~al.(2017)Howard, Zhu, Chen, Kalenichenko, Wang, Weyand, Andreetto, and Adam]{howard2017mobilenets}
A.~G. Howard, M.~Zhu, B.~Chen, D.~Kalenichenko, W.~Wang, T.~Weyand, M.~Andreetto, and H.~Adam, ``Mobilenets: Efficient convolutional neural networks for mobile vision applications,'' 2017.

\bibitem[Tian et~al.(2019)Tian, Shen, Chen, and He]{tian2019fcos}
Z.~Tian, C.~Shen, H.~Chen, and T.~He, ``Fcos: Fully convolutional one-stage object detection.'' 2019.

\bibitem[Li and Ren(2019)]{li2019light}
Y.~Li and F.~Ren, ``Light-weight retinanet for object detection,'' \emph{arXiv preprint arXiv:1905.10011}, 2019.

\bibitem[Liu et~al.(2016)Liu, Anguelov, Erhan, Szegedy, Reed, Fu, and Berg]{liu2016ssd}
W.~Liu, D.~Anguelov, D.~Erhan, C.~Szegedy, S.~Reed, C.-Y. Fu, and A.~C. Berg, ``Ssd: Single shot multibox detector,'' in \emph{ECCV 2016: 14th European Conference}.\hskip 1em plus 0.5em minus 0.4em\relax Springer, 2016.

\bibitem[Simonyan(2014)]{simonyan2014very}
K.~Simonyan, ``Very deep convolutional networks for large-scale image recognition,'' 2014.

\end{thebibliography}

\end{document}